\documentclass[10pt,twocolumn,letterpaper]{article}

\usepackage[pagenumbers]{cvpr} 

\usepackage[dvipsnames]{xcolor}
\newcommand{\red}[1]{{\color{red}#1}}

\usepackage{graphicx}
\usepackage{amsmath}
\usepackage{amssymb}
\usepackage{booktabs}

\definecolor{cvprblue}{rgb}{0.21,0.49,0.74}
\usepackage[pagebackref,breaklinks,colorlinks,citecolor=cvprblue]{hyperref}

\usepackage[capitalize]{cleveref}
\crefname{section}{Sec.}{Secs.}
\Crefname{section}{Section}{Sections}
\Crefname{table}{Table}{Tables}
\crefname{table}{Tab.}{Tabs.}

\def\confName{CVPR}
\def\confYear{2024}

\usepackage{color}
\usepackage{xcolor}
\definecolor{deepGreen}{RGB}{0,153,0}
\definecolor{orange}{RGB}{255,125,0}
\def\red#1{\textcolor[rgb]{1,0,0}{#1}}

\def\graytext#1{\textcolor[RGB]{85,85,85}{#1}}
\def\GTgreen#1{\textcolor[RGB]{0,204,0}{#1}}

\definecolor{sainone}{RGB}{236, 242, 249}
\definecolor{saintwo}{RGB}{255, 230, 204}

\newcommand{\keypoint}[1]{\vspace{0.1cm}\noindent\textbf{#1}\;}
\newcommand{\cut}[1]{}
\usepackage{multirow}
\usepackage{amsmath}
\usepackage{amssymb}
\usepackage{mathtools}
\usepackage{arydshln}
\usepackage{soul}
\usepackage{nicefrac}
\usepackage{booktabs}
\usepackage{stmaryrd}
\usepackage{caption}
\usepackage{graphicx}
\usepackage{colortbl}
\definecolor{gray}{gray}{0.9}
\definecolor{pink}{RGB}{255, 234, 232}
\sethlcolor{pink}
\usepackage[accsupp]{axessibility}

\newcommand{\MYhref}[3][blue]{\href{#2}{\color{#1}{#3}}}

\makeatletter
\apptocmd\@maketitle{{\myfigure{}\par}}{}{}
\makeatother
\usepackage{capt-of,etoolbox}

\makeatletter
\newcommand\notsotiny{\@setfontsize\notsotiny\@vipt\@viipt}
\makeatother

\usepackage{xcolor,pifont}
\newcommand*\colourcheck[1]{%
  \expandafter\newcommand\csname #1check\endcsname{\textcolor{#1}{\ding{52}}}%
}
\newcommand*\colourcross[1]{%
  \expandafter\newcommand\csname #1cross\endcsname{\textcolor{#1}{\ding{55}}}%
}
\colourcheck{blue}
\colourcheck{green}
\colourcross{red}

\title{\vspace{-1cm}You'll Never Walk Alone: A Sketch and Text Duet for \\ Fine-Grained Image Retrieval \vspace{-0.75cm}}

\author{\MYhref[cvprblue]{https://subhadeepkoley.github.io}{Subhadeep Koley}\textsuperscript{1,2} \hspace{.2cm} \MYhref[cvprblue]{https://ayankumarbhunia.github.io}{Ayan Kumar Bhunia}\textsuperscript{1} \hspace{.2cm} \MYhref[cvprblue]{https://aneeshan95.github.io}{Aneeshan Sain}\textsuperscript{1} \hspace{.2cm}  \MYhref[cvprblue]{https://www.pinakinathc.me}{Pinaki Nath Chowdhury}\textsuperscript{1} \\ \MYhref[cvprblue]{https://www.surrey.ac.uk/people/tao-xiang}{Tao Xiang}\textsuperscript{1,2} \hspace{.2cm} \MYhref[cvprblue]{https://www.surrey.ac.uk/people/yi-zhe-song}{Yi-Zhe Song}\textsuperscript{1,2} \\
\textsuperscript{1}SketchX, CVSSP, University of Surrey, United Kingdom.  \\
\textsuperscript{2}iFlyTek-Surrey Joint Research Centre on Artificial Intelligence.\\
{\tt\small \{s.koley, a.bhunia, a.sain, p.chowdhury, t.xiang, y.song\}@surrey.ac.uk}\\
\small \url{https://subhadeepkoley.github.io/Sketch2Word}
}
\vspace{-0.5cm}

\newcommand\myfigure{
\centering
\vspace{-0.9cm}
\captionsetup{type=figure} 
\setlength{\tabcolsep}{0.9pt}
  \renewcommand{\arraystretch}{1.1}
  \scriptsize
    \begin{minipage}{1\linewidth}
    \centering
    \includegraphics[width=1\linewidth]{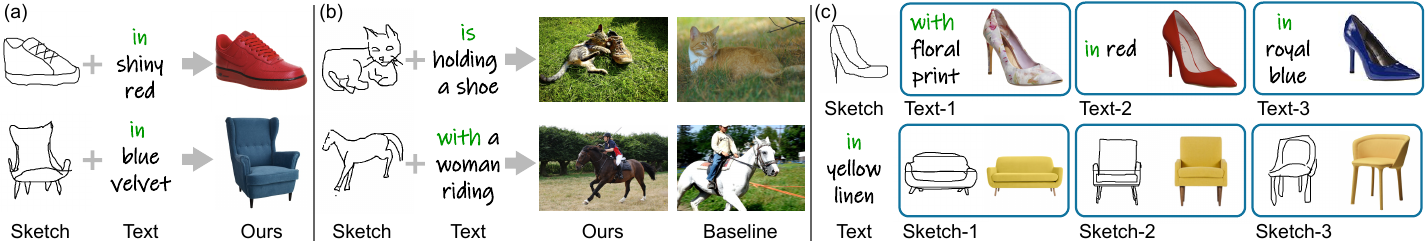}
  \end{minipage}
  \vspace{-0.3cm}
    \captionof{figure}{
    (a) Photos retrieved by our method, depicting precise control over \textit{both} shape and appearance. (b) Unlike baseline sketch+text composed retrieval framework, our method seamlessly composes the \textit{structural} and \textit{contextual} cues of sketch and text queries respectively. (c) With a \textit{fixed} sketch query, our method retrieves \textit{different} images for \textit{different} textual descriptions and vice-versa, depicting the \textit{complementarity} of sketch and text modalities in sketch+text-based composed image retrieval. For a fixed sketch, the visual \textit{attributes} from different textual descriptions are visibly reflected in the retrieved images while maintaining shape consistency. Similarly, fixing the attributes provided via text, shapes of retrieved images change corresponding to different sketch queries.
    }
\label{fig:teaser}
\vspace{+0.1cm}
}

\begin{document}
\maketitle

\begin{abstract}
\vspace{-0.3cm}

Two primary input modalities prevail in image retrieval: sketch and text. While text is widely used for inter-category retrieval tasks, sketches have been established as the sole preferred modality for fine-grained image retrieval due to their ability to capture intricate visual details. In this paper, we question the reliance on sketches alone for fine-grained image retrieval by simultaneously exploring the fine-grained representation capabilities of both sketch and text, orchestrating a duet between the two. The end result enables precise retrievals previously unattainable, allowing users to pose ever-finer queries and incorporate attributes like colour and contextual cues from text. For this purpose, we introduce a novel compositionality framework, effectively combining sketches and text using pre-trained CLIP models, while eliminating the need for extensive fine-grained textual descriptions. Last but not least, our system extends to novel applications in composed image retrieval, domain attribute transfer, and fine-grained generation, providing solutions for various real-world scenarios.

\end{abstract}

\vspace{-0.6cm}
\section{Introduction}
\vspace{-0.1cm}
Sketch and text represent the two most common~\cite{chowdhury2023scenetrilogy, sangkloy2022sketch} input modalities in the realm of image retrieval. The choice between these modalities depends on the nature of the retrieval problem, especially when fine-grained distinctions are required~\cite{yelamarthi2018zero, song2017fine, dutta2019semantically, sangkloy2022sketch}. In inter-category retrieval, text dominates as the primary modality, exemplified by widely-used platforms like Google Images. However, when the challenge transitions to fine-grained image retrieval, sketches take the spotlight~\cite{chowdhury2023scenetrilogy, song2017fine, sangkloy2022sketch}. Sketches promise to capture fine-grained visual cues that can be cumbersome or even impossible for text to express~\cite{chowdhury2023scenetrilogy}. Research in this domain predominantly revolves around harnessing the unique qualities of sketches, exploring aspects such as style~\cite{sain2021stylemeup}, abstraction~\cite{koley2024how}, and more~\cite{dutta2019semantically, bhunia2022sketching, sangkloy2022sketch}.

In this paper, we question this notion that ``sketch is everything'' and, for the first time, simultaneously delve into the fine-grained representation capabilities of both sketch and text, and in turn orchestrate a duet between these two modalities for fine-grained image retrieval. The outcome is a novel retrieval experience where a sketch and text work in harmony, enabling users to achieve precise retrievals that were previously unattainable. Now, users can locate ``that'' specific shoe, considering not only fine-grained pattern cues from sketches but also incorporating attributes like colour and texture from text (\cref{fig:teaser}a). This synergy extends to scenarios where text offers contextual cues to a given sketch, such as a cat holding a shoe for that matter (\cref{fig:teaser}b)!

While sketch and text synergy has been studied before~\cite{chowdhury2023scenetrilogy, sangkloy2022sketch}, it has predominantly focused on \textit{scene-level/category} retrieval, where paired sketch and text descriptions for a given {scene/category} are readily available in datasets (\eg, FS-COCO~\cite{chowdhury2022fs}, CM-Places~\cite{castrejon2016learning}). Our contention is that the synergy between sketches and text, while notable, is not as pronounced in category-level retrieval as it is for \textit{fine-grained} retrieval~\cite{chowdhury2023scenetrilogy,castrejon2016learning}. Indeed, for category-level retrieval, one might argue the necessity of sketches, as the descriptive power of text might already suffice~\cite{sangkloy2022sketch}. However, when the desire extends to a horse/cat with a specific {pose}, an accompanying sketch becomes indispensable (as illustrated in \cref{fig:teaser}b).

Our foremost challenge centres around fine-grained compositionality, specifically investigating how sketches and text can serve as complementary components in fine-grained queries. Our goal is to maintain the semantics of both modalities, ensuring, for example, that a horse in \cref{fig:teaser}b corresponds precisely to the respective sketch and associated text, rather than any generic horse with a woman riding. To tackle this challenge, we harness the capabilities of CLIP \cite{radford2021learning}, leveraging its implicit grammatical-composition capability. We achieve this via CLIP \cite{radford2021learning} to create a \textit{fine-grained textual equivalent} of the input sketch, referred to as a ``pseudo-word token''. This token, when combined with text input, forms a fine-grained textual query that seamlessly integrates both sketch and text features, allowing them to work in synergy within the text domain.

The second challenge pertains to alleviating the requirement of collecting a dataset of fine-grained sketch and text pairs. We also aim to emulate the inference-time distribution of text input. The key innovation lies in the hypothesis that the fine-grained description embedded in a photo (P) can be approximated by that of a sketch (S) plus text (T), leading to T = P - S. This relationship illustrates how the absence of text can be approximated by the difference signal between the photo and the sketch in the latent embedding space. We incorporate this difference signal as a \textit{proxy} for the missing textual description during training to make it inference-time equivalent through a novel \textit{compositionality constraint}. Furthermore, we utilise short phrases generated by a lightweight GPT~\cite{brown2020language} as a \textit{neutral-text regulariser} to ensure that the synergy works without disrupting the grammatical structure of CLIP's \cite{radford2021learning} language manifold.

Last but not least, in addition to addressing the challenges in fine-grained image retrieval, our system opens the door to a range of novel applications in the field of composed image retrieval such as object-sketch-based scene image retrieval, domain attribute transfer, and sketch+text-based fine-grained image generation.

In summary, \textit{(i)} we address the challenge of fine-grained image retrieval by leveraging the synergy between freehand sketches and textual descriptions, extending retrieval beyond traditional category-level distinctions. \textit{(ii)} we introduce a novel compositionality framework, effectively combining sketches and text using pre-trained CLIP models, eliminating the need for extensive fine-grained textual descriptions. \textit{(iii)} our system unlocks novel applications like object-sketch-based scene retrieval, domain attribute transfer, and sketch+text-based fine-grained image generation.

\vspace{-0.3cm}
\section{Related Works}
\vspace{-0.2cm}

\noindent \textbf{Sketch-Based Image Retrieval (SBIR).} Starting at category-level, SBIR is tasked to fetch a photo of the same category as that of a given query sketch. Earlier deep-learning methods~\cite{liu2017deep, yang2020deep, zhang2019learning, collomosse2019livesketch} generally train Siamese-like networks~\cite{collomosse2019livesketch} over a distance-metric in a cross-modal joint embedding space~\cite{collomosse2017sketching}. Moving forward to \textit{fine-grained} SBIR (FG-SBIR), the aim is to retrieve one particular \textit{photo-instance} from a gallery of same-category photos corresponding to a query sketch. FG-SBIR has progressed from a deep-triplet ranking-based Siamese network~\cite{yu2016sketch} to further enhancements involving higher-order attention~\cite{song2017deep} or auxiliary losses~\cite{lin2019tc}, and local feature alignment~\cite{xu2021dla}, to name a few. While most works focus on various applications like early-retrieval~\cite{bhunia2020sketch}, cross-category generalisation~\cite{pang2019generalising, bhunia2022adaptive} and even zero-shot FG-SBIR~\cite{sain2023clip, koley2024text}, others delved deeper to explore sketch-specific traits, like hierarchy~\cite{sain2020cross}, or style-diversity~\cite{sain2021stylemeup}, for better retrieval. Recent extensions include retrieving a scene image based on a scene-sketch (\ie, Scene-level FG-SBIR), employing cross-modal region associativity~\cite{chowdhury2022partially}, enhanced further with text-query~\cite{chowdhury2023scenetrilogy}. Unlike existing \textit{fully-supervised} sketch+text-based methods~\cite{chowdhury2023scenetrilogy, sangkloy2022sketch}, relying on \textit{paired} training triplets (\ie, sketch, text, and ground truth photo), our approach alleviates the need for such triplets, simplifying the challenge of collecting fine-grained sketch-text-photo dataset.

\keypoint{Text-Based Image Retrieval (TBIR).} Over the years, much emphasis has been paid to textual query-based image retrieval by learning a joint embedding space via ranking loss~\cite{eisenschtat2017linking, karpathy2015deep, plummer2018conditional}. This was further augmented by cross-modal message passing~\cite{wang2019camp}, hard triplet mining~\cite{faghri2017vse++}, and
one-to-many probabilistic mapping~\cite{chun2021probabilistic}, to name a few. Thanks to internet-scale paired image-text datasets TBIR has become highly competitive and one of the most active areas of research~\cite{radford2021learning}, leading to expansive techniques like Oscar~\cite{li2020oscar}, CLIP~\cite{radford2021learning}, ALIGN~\cite{jia2021scaling}, etc. The recent paradigm of \textit{Textual Inversion}~\cite{cohen2022my, gal2022image} deals with inverting input image(s) into a pseudo-word token in the language space of pre-trained vision-language models for downstream tasks like personalised image retrieval~\cite{cohen2022my}, composed retrieval~\cite{saito2023pic2word, baldrati2023zero}, etc. Composed image retrieval (CIR) aims to retrieve images from a combined query of text and image pairs~\cite{baldrati2022effective}. Existing CIR methods typically leverage pre-trained CLIP~\cite{baldrati2022effective, baldrati2023zero, saito2023pic2word}, or resort to image-text feature fusion~\cite{han2022fashionvil, goenka2022fashionvlp, wu2021fashion, vo2019composing}. Textual inversion-based CIR methods~\cite{baldrati2023zero, saito2023pic2word} either use million-scale image dataset to train inversion networks~\cite{saito2023pic2word}, or uses time-consuming two-stage optimisation-based approach~\cite{baldrati2023zero}. Nevertheless, these \textit{image}+text composed retrieval frameworks~\cite{baldrati2023zero, saito2023pic2word} can not handle the huge domain gap of \textit{sparse sketch}+text composed image retrieval. Our method, on the other hand, explicitly models the sketch-photo \textit{difference} within itself for fine-grained sketch+text composed multi-modal retrieval.

\keypoint{Sketch+Text Joint Multi-Modal Learning.} While sketch is a suitable ~\cite{chowdhury2022fs} fine-grained query medium for AI systems, certain aspects fall outside its scope like qualitative attributes (\eg, colour, shade, etc.)~\cite{chowdhury2023scenetrilogy}. Text aptly describes these qualitative attributes (\eg, colour) which leads to ``text as query" being extensively studied~\cite{chowdhury2023scenetrilogy}, albeit largely for category-level tasks (\eg, TBIR). Realising this potential of sketch-text-photo association, joint multi-modal sketch+text query was heavily used in downstream vision tasks. On generation, sketch-to-image models used text-as-guidance to specify class-label \cite{ham2022cogs}. Simultaneously text-to-image diffusion models used sketches as semantic-guidance for the encoder \cite{t2i-adapter}, or decoder \cite{controlNet}, or external classifier guidance \cite{voynov2023sketch}, NeRF-editing \cite{mildenhall2021nerf, muller2022instant} using sketch+text conditioning \cite{mikaeili2023sked}, and sketch-conditioned image captioning \cite{chowdhury2023scenetrilogy}. On retrieval, the importance of sketch+text was realised with FS-COCO \cite{chowdhury2022fs} dataset collecting scene-level sketch-text-photo dataset. Recent works include supervised approaches of Sangkloy \etal \cite{sangkloy2022sketch} using CLIP, and Scenetrilogy \cite{chowdhury2023scenetrilogy} using conditional invertible neural networks for sketch+text-based image retrieval. Despite its benefit, collecting fine-grained textual descriptions is quite cumbersome~\cite{chowdhury2022fs}, which bottlenecks further exploration of fine-grained sketch-text-photo association for downstream tasks to date.

\vspace{-0.3cm}
\section{Revisiting CLIP}
\vspace{-0.2cm}
CLIP~\cite{radford2021learning} consists of a text encoder and an image encoder, trained with a multi-class N-pair contrastive loss~\cite{radford2021learning} on internet-scale ($\sim$$400M$) image-text pair dataset, with an aim to learn a joint-embedding space that minimises the cosine similarity between the matching image-text pairs while maximising the same for random unpaired ones~\cite{radford2021learning}. The image encoder ($\mathbf{V}$) usually employs a Vision Transformer (ViT) \cite{dosovitskiy2021image}, to encode an input image $\mathcal{I}$ into a visual feature as $\mathbf{i} = \mathbf{V}(\mathcal{I}) \in \mathbb{R}^d$. The text encoder ($\mathbf{T}$) inputs a sequence of words $\mathcal{W} = \{w^0, w^1, \dots, w^k\}$ and applies a lower-cased byte pair encoding (BPE), followed by a learnable word embedding layer $\mathbf{T_w}$ ($49,152$ vocab size) \cite{radford2021learning}, to convert each word $w^i$ into a \emph{word token embedding} $\mathbf{w}_e^i = \mathbf{T_w}(w^i)$ of size $\mathbb{R}^d$. This sequence of {word token embeddings} $ \mathcal{W}_e = \{\mathbf{w}^0_e, \mathbf{w}^1_e, \dots, \mathbf{w}^k_e\}$ is then passed via a transformer $\mathbf{T_t}$, to provide the final textual feature as $\mathbf{w} = \mathbf{T_t}(\mathcal{W}_e) \in \mathbb{R}^d$, taken from the last hidden state of the final transformer layer.

\begin{figure*}[!htbp]
    \centering
    \includegraphics[width=\textwidth]{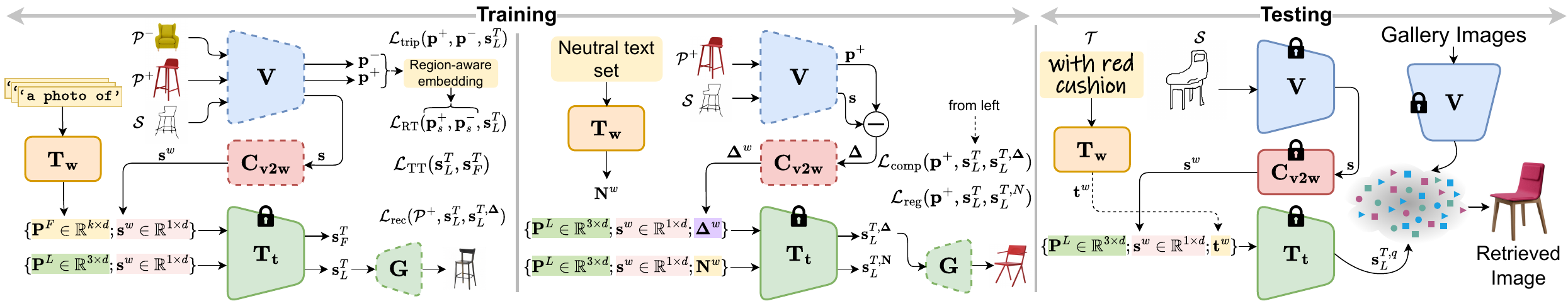}
    \vspace{-0.7cm}
    \caption{
    A query sketch $(\mathcal{S})$ is passed via CLIP's visual encoder $(\mathbf{V})$ followed by the visual-to-word converter $(\mathbf{C_{v2w}})$ to obtain pseudo-word token embedding $(\mathbf{s}^w)$. It is then appended with a learnable continuous prompt $\mathbf{P}^L \in \mathbb{R}^{3\times d}$ and passed via frozen $\mathbf{T_t}$ to produce the final sketch embedding $\mathbf{s}_L^T$. \textit{Compositionality constraint} (middle) is importantly a part of our multitask training (\textit{not} a two-stage approach \cite{baldrati2023zero, saito2023pic2word}), where we compute $\mathbf{s}_L^{T,\Delta}$ (\cref{composition}) by passing the \emph{sketch-photo difference signal} $\mathbf{\Delta}$ via $\mathbf{C_{v2w}}$ and appending as $\mathbf{s}_L^{T,\Delta}$=$\{\mathbf{P}^L; \mathbf{s}^w; \mathbf{\Delta}^w\}$, using which $\mathcal{L}_{\text{comp}}$ is imposed. However, as this numeric signal $\mathbf{\Delta}^w$ does not exist in CLIP's \cite{radford2021learning} input text manifold, it may disrupt its grammatical syntax. Thus, we mine a set of ``neutral text'' (via GPT \cite{brown2020language}) to impose a regularisation loss $\mathcal{L}_{\text{reg}}$. Apart from $\mathcal{L}_{\text{trip}}$, we use $\mathcal{L}_{\text{RT}}$ (region-aware triplet) with $\mathbf{s}_L^T$ and photo embeddings $\mathbf{p^+}/\mathbf{p^-}$ to enforce fine-grained matching. Additionally, a reconstruction loss $\mathcal{L}_{\text{rec}}$ trains a UNet decoder $(\mathbf{G})$ for further cross-modal alignment (\cref{fine_grained}). Furthermore, $\mathcal{L}_{\text{TT}}$, using a pre-defined set of standard language prompts, brings learnable prompts closer to \textit{actual} English prompts for \textit{unseen set} generalisation (\cref{text_text}). Specifically, we only train the $\mathtt{LayerNorm}$ of $\mathbf{V}$, $\mathbf{C_{v2w}}$, $\mathbf{P}^L$, and $\mathbf{G}$. The testing pipeline is shown on the right. \textit{(Best view when zoomed.)}}
    \label{fig:framework}
    \vspace{-0.4cm}
\end{figure*}

\vspace{-0.3cm}
\section{Sketch-Based Composed Image Retrieval}
\vspace{-0.2cm}

\keypoint{Motivation.}
Combining structural cues from \textit{sketch} with additional \textit{textual description} results in a powerful query for image retrieval. Existing works \cite{sangkloy2022sketch, chowdhury2023scenetrilogy} on such compositionality usually extract sketch and text features via separate encoders, and add or concatenate (followed by additional learnable layers) them, to obtain the composed query feature. This has two major issues: \textit{(i)} it needs sketch-\textit{associated} textual description -- absent from fine-grained SBIR datasets \cite{yu2016sketch, sangkloy2016the}, and \textit{(ii)} combining the two features naively may distort the optimal sketch-text feature correlation needed, to correctly represent a composed semantic. This \textit{compositionality} is however more explicit in the textual domain \cite{baldrati2023zero, cohen2022my, saito2023pic2word} where combining individual words/phrases form a composed semantic, \eg, {`a cat'} and {`brown'} together infers \textit{`a brown cat'}. Following CLIP's rise in various downstream tasks \cite{khattak2023maple, zhou2022conditional, sain2023clip}, we thus aim to leverage its text encoder's input text space to tackle sketch+text compositionality. In particular, {inspired by textual inversion literature \cite{gal2022image},} we aim to represent a sketch as a \textit{pseudo-word token} that emulates its visual concept in equivalent word-embedding space, and combine its textual description via connecting phrases like `with', `in', `and', etc.\ (the full list in \red{\S}~Suppl.) to obtain ``$\langle$pseudo-word token$\rangle$ $\langle$connecting phrase$\rangle$ $\langle$text description$\rangle$" as a query. Passing this via CLIP's text encoder would provide a sketch+text composed representation, that \textit{can} be compared against gallery image features pre-computed via CLIP's vision encoder. The goal here is to learn sketch+text compositionality via CLIP, unsupervised, without any expensive paired textual description.

\keypoint{Overall Framework.} Here, we aim to design a Sketch-Based Composed Image Retrieval (SBCIR) framework harnessing the Vision-Language (V-L) embedding of pre-trained CLIP~\cite{radford2021learning} using only the sketch-photo pairing readily available in sketch datasets \cite{yu2016sketch, sangkloy2016the, chowdhury2022fs}, \emph{without} any annotated textual description. Accordingly, we embed the sketch into a pseudo-word token by first passing it through CLIP's visual encoder $\mathbf{V}$, followed by a visual-to-word converter $\mathbf{C_{v2w}}$, which is then passed into CLIP's text transformer $\mathbf{T_t}$ along with a few learnable prompts and additional textual descriptions (during inference) to obtain the composed query embedding. Specifically, we have three salient designs -- \textit{(i)} a novel compositionality constraint imposed via sketch/photo difference-signal to imitate the missing textual description (during training) and \textit{neutral text} (\cref{composition}) to preserve the grammatical structure of the input text-space of CLIP's text encoder, \textit{(ii)} generalisable continuous prompt-learning (\cref{text_text}) over handcrafted textual prompts,  and \textit{(iii)} fine-grained matching (\cref{fine_grained}) between composed query and paired photo embedding via region-aware triplet loss and an auxiliary generative loss.

\subsection{Baseline SBCIR}
\label{baseline}

Conventional CLIP-based SBIR~\cite{sain2023clip} maps both sketch query and its target photo in the joint embedding space using the visual encoder ($\mathbf{V}$). On the contrary, we convert a query sketch to a \textit{pseudo-word token} and pass it through CLIP's textual encoder ($\mathbf{T_t}$) to generate its representation. In particular, given a photo $\mathcal{P}$, we generate its latent embedding ($\mathbf{p}$) as $\mathbf{p} = \mathbf{V}(\mathcal{P})\in\mathbb{R}^{1\times d}$. For a sketch however, we generate its equivalent word embedding as $\mathbf{s}^w=\mathbf{C_{v2w}}(\mathbf{V}(\mathcal{S}))\in\mathbb{R}^{1\times d}$. $\mathbf{s}^w$ signifies the equivalent language representation of the visual sketch query. We design the visual-to-word converter network $\mathbf{C_{v2w}}$ using 3-Layer MLP with ReLU~\cite{nair2010rectified}. Inspired from prompt learning~\cite{zhou2022conditional, zhou2022learning}, we prepend a $\mathbb{R}^{3\times d}$ learnable continuous prompt vector $\mathbf{P}^L$ with $\mathbf{s}^w$ and pass this through $\mathbf{T_t}$ to finally obtain $\mathbf{s}^T_L = \mathbf{T_t}(\{\mathbf{P}^L;\mathbf{s}^w\}) \in\mathbb{R}^{1\times d}$ . Baseline SBCIR trains over triplet loss~\cite{weinberger2009distance} to perform cross-modal matching between $\mathbf{s}^T_L$, and positive ($\mathbf{p}^+$) and negative ($\mathbf{p}^-$) photo features. With margin $\mu_{\text{trip}}>0$, triplet loss aims to minimise the distance $\delta(\cdot,\cdot)$ between $\mathbf{s}^T_L$ and $\mathbf{p}^+$, while increasing that from a random negative feature $\mathbf{p}^-$.
\begin{equation}
    \mathcal{L}_{\text{trip}}=\mathtt{max}(0,\mu_{\text{trip}}+\delta(\mathbf{s}^T_L, \mathbf{p}^+)-\delta(\mathbf{s}^T_L, \mathbf{p}^-))
    \label{vanila_trip}
\end{equation}

We only train the $\mathtt{LayerNorm}$ parameters of $\mathbf{V}$ \cite{sain2023clip}, $\mathbf{C_{v2w}}$, and $\mathbf{P}^L$, while the rest including $\mathbf{T_t}$ remains frozen. Consequently, we can leverage the zero-shot compositionality of CLIP's text encoder $\mathbf{T}$ to perform SBCIR even if \textit{no} text was provided during training. During inference, we append the word embedding of optional textual query ($\mathcal{T}$) as $\mathbf{t}^w = \mathbf{T_w}(\mathcal{T})$ to form $\{\mathbf{P}^L;\mathbf{s}^w; \mathbf{t}^w\}$, and pass it through frozen $\mathbf{T_t}$ to get the final composed query vector.

However, baseline SBCIR has a few major limitations -- \textit{(i)} most importantly, without access to paired textual descriptions during training, it \textit{solely} relies on the frozen CLIP encoder~\cite{radford2021learning} for zero-shot compositionality. We thus set out to explore how we can \textit{imitate} the effect of adding this query text during training to learn the compositionality, \textit{(ii)} while CLIP~\cite{radford2021learning} performs fairly well for \textit{category-level} semantic matching~\cite{sain2023clip, sangkloy2022sketch}, its \textit{off-the-shelf} adaptation in the \textit{fine-grained} setting is sub-optimal as seen in certain CLIP-based works~\cite{luo2023segclip,sain2023clip}. So, how can we make the cross-modal matching more \textit{fine-grained}? \textit{(iii)} prompt learning is prone to overfitting on the training set, delivering poor test-set zero-shot performance~\cite{bulat2023lasp}. Furthermore, vanilla prompt learning is not robust~\cite{bulat2023lasp} and is prone to distort the composed query embedding, particularly when combined with the optional text coming from an \textit{in-the-wild} scenario.

\subsection{Learning Compositionality Constraint}
\label{composition}
While our method does not rely on paired textual descriptions \textit{during training}, users can provide optional textual descriptions \textit{during inference}, to specify the desired \textit{additional information}, which gets composed with the query sketch to retrieve the correct image. We handle this training-testing disparity, by hypothesising the \textit{additional information} to be equivalent to the difference between the query sketch and its paired photo. To this end, we compute the sketch-photo \textit{difference} signal embedding $\mathbf{\Delta}^w=\mathbf{C_{v2w}(|\mathbf{p}^+-\mathbf{s}|)}$ and append it as $\{\mathbf{P}^L;\mathbf{s}^w;\mathbf{\Delta}^w\}$, which upon passing through $\mathbf{T_t}$ generates the $\mathbf{s}_L^{T,\mathbf{\Delta}}$ embedding. Here $\mathbf{\Delta}^w$ could be considered as a ``single vector'' pseudo-word token \textit{imitating} the difference between sketch and photo, which ideally would be substituted with real query text during inference. Now, as per our hypothesis, we enforce compositionality constraint $\mathcal{L}_{\text{comp}}$ (with $\mu_{\text{comp}}>0$) that ensures that the distance between $\mathbf{p}^+$ and $\mathbf{s}_L^{T,\mathbf{\Delta}}$ is less than the same between $\mathbf{p}^+$ and $\mathbf{s}_L^{T}$, as $\mathbf{\Delta}^w$ reinforces $\mathbf{s}_L^{T}$ with \textit{additional information}.

\vspace{-0.5cm}
\begin{equation}
    \mathcal{L}_{\text{comp}}=\mathtt{max}(0,\mu_{\text{comp}}+\delta(\mathbf{s}^{T,\mathbf{\Delta}}_{L}, \mathbf{p}^+)-\delta(\mathbf{s}^{T}_{L}, \mathbf{p}^+))
    \vspace{-0.1cm}
\end{equation}

Although $\mathbf{\Delta}$ enforces compositionality, this mere numeric signal does not exist in CLIP's~\cite{radford2021learning} input text manifold and might break its grammatical syntax. Furthermore, this additional signal might train $\mathbf{C_{v2w}}$ sub-optimally, rendering its output incompatible with CLIP's input text manifold. Thus, to restrict the adverse effect of $\mathbf{\Delta}$ to a minimum level, we regularise the training via a \textit{``neutral-text''} set containing a list of $3$-$5$ word \emph{generic} description of a freehand sketch (\eg, $\mathtt{``with~a~line~drawing"}$). To this end, we replace $\mathbf{\Delta}^w$ from composed query $\mathbf{s}_L^{T,\mathbf{\Delta}}$ with any one random phrase from the neutral-text set, to generate neutral-text enriched composed representation $\{\mathbf{P}^L;\mathbf{s}^w;\mathbf{N}^w\}$, which upon passing through $\mathbf{T_t}$ generates the $\mathbf{s}_L^{T,N}$ embedding. Here $\mathbf{N}^w$ is the word-embedding of the chosen neutral phrase. We posit that using a \textit{generic description} for a sketch should neither enhance nor impair the composed query. Thus, we enforce the distance between $\mathbf{p}^+$ and $\mathbf{s}_L^{T,\mathbf{\Delta}}$ to be \textit{equivalent} to the distance between $\mathbf{p}^+$ and $\mathbf{s}_L^{T,N}$.

\vspace{-0.2cm}
\begin{equation}
    \mathcal{L}_{\text{reg}}={||\delta(\mathbf{s}^{T,\mathbf{\Delta}}_{L}, \mathbf{p}^+)-\delta(\mathbf{s}^{T,N}_{L}, \mathbf{p}^+)||}_2
\end{equation}

We prompt a lightweight GPT~\cite{brown2020language}, to generate $100$ different $3$-$5$ word phrases (more in \red{\S}~Suppl.) describing a ``freehand sketch'' to form our optimum neutral text set.

\subsection{Generalised Prompt Learning}
\label{text_text}
Prompt learning literature~\cite{zhou2022learning, zhou2022conditional, bulat2023lasp} dictates that handcrafted fixed language prompts (\eg, $\mathtt{``a~photo~of"}$, $\mathtt{``an~image~of"})$ generalise better on \textit{unseen} sets, while learnable continuous prompts depict better performance on \textit{seen} sets used to learn it. While we are using learnable continuous prompts $\mathbf{P}^L$ over handcrafted ones, we impose a \textit{text-to-text} generalisation loss that enforces the learned prompts to be similar to a set of handcrafted language prompts in the text embedding space, so that it generalises beyond the seen training set. In particular, at every instance, we randomly pick one handcrafted fixed language prompt $d_i$ from a set of handcrafted \cite{zhou2022learning} fixed language prompts (more in \red{\S}~Suppl.) $\mathcal{D}$ and prepend its word-embedding representation $\mathbf{P}^F = \mathbf{T_w}(d_i)$ to $\mathbf{s}^w$ as $\{\mathbf{P}^F,\mathbf{s}^w\}$ and pass it via $\mathbf{T_t}$ to generate the fixed representation $\mathbf{s}^T_F$. Now we employ $\mathcal{L}_\text{TT}$ to enforce the sketch query representation with the learned prompts $\mathbf{s}^T_L$ to be similar with that of the fixed language prompt $\mathbf{s}^T_F$ as:

\vspace{-0.3cm}
\begin{equation}
    \mathcal{L}_\text{TT} = {||\mathbf{s}^T_F - \mathbf{s}^T_L||}_2
    \vspace{-0.05cm}
\end{equation}

The utility of $\mathcal{L}_\text{TT}$ is multi-fold -- \textit{(i)} it alleviates seen set overfitting, \textit{(ii)} typically, learned prompts reside in the sparse regions of the CLIP manifold~\cite{baldrati2023zero}, limiting its intractability with actual query texts during inference. Regularising $\mathbf{P}^L$ with actual language supervision suppresses this issue, and \textit{(iii)} the diverse list of fixed prompts, acts as an additional augmentation in the language domain~\cite{bulat2023lasp}, which reinforces the robustness of the learned prompts.

\vspace{-0.1cm}
\subsection{Fine-Grained Matching}
\vspace{-0.2cm}
\label{fine_grained}
\keypoint{Region-Aware Triplet Loss.} To further improve the cross-modal fine-grained matching, we consider the CLIP~\cite{radford2021learning} vision encoder $\mathbf{V}$ (employed through a vision transformer) that breaks the input image ($\mathcal{P}$) into $T$ patches and passes them via transformer layer to generate patch-wise feature $\mathbf{p}_r = \mathbf{V}(\mathcal{P})\in\mathbb{R}^{T\times d}$, where $T$ is the number of patches. To enforce region-wise associativity, we use the patch-wise embedding $\mathbf{p}_r$ from all $T$ patches and calculate the patch-level correlation ($a$) between global sketch query feature $\mathbf{s}_L^T\in\mathbb{R}^{1\times d}$ and $\mathbf{p}_r$ as:  $a=(\mathbf{p}_r \cdot \mathbf{s}_L^T) \in\mathbb{R}^{T\times 1}$, which is $\mathtt{SoftMax}$ normalised across the patch dimension. Every value $a_i$ denotes the associativity between the global sketch query and patch-wise photo features. Now we take a weighted sum across all patch embeddings to get a region-aware photo feature as: ${\mathbf{p}}_s = \sum_{i=1}^{T}(a_i\times \mathbf{p}_r^i)$. We utilise this \textit{region-aware embeddings} $\mathbf{p}^+_s$ and $\mathbf{p}^-_s$ from the positive and the negative photo respectively to impose one additional region-aware triplet loss $\mathcal{L}_{\text{RT}}$ with margin $\mu_\text{RT}>0$ as:

\vspace{-0.6cm}
\begin{equation}
    \mathcal{L}_\text{RT} = \mathtt{max}(0,\mu_{\text{RT}}+\delta(\mathbf{s}^T_L, \mathbf{p}^+_s)-\delta(\mathbf{s}^T_L, \mathbf{p}^-_s))
    \label{region_trip}
    \vspace{-0.2cm}
\end{equation}

It is noteworthy that although the $\mathcal{L}_{\text{RT}}$ acts as a proxy to better align the joint embedding space for fine-grained matching, we perform inference on the \textit{global feature}.

\keypoint{Auxiliary Generator Guidance.} With sketches, triplet loss typically performs fine-grained shape matching~\cite{yu2016sketch, bhunia2020sketch}, ignoring the \textit{fine-grained} appearance features (\eg, colour, texture). Being a \textit{composed} retrieval framework, we aim to encompass appearance traits along with structural ones
in the visual domain. Considering the proven efficacy of cross-modal translation~\cite{pang2017cross} in fine-grained matching, we impose a sketch-to-photo reconstruction objective, where given the sketch query representation, we train a simple UNet~\cite{ronneberger2015u} decoder ($\mathbf{G}$) to reconstruct the ground truth photo using pixel-level $l_2$ reconstruction loss. Please note, that the aim of $\mathcal{L}_{\text{rec}}$ is not to enforce photorealistic image generation, but to impose an appearance guidance.

\vspace{-0.5cm}
\begin{equation}
    \mathcal{L}_{\text{rec}} = {||\mathcal{P}^+-\mathbf{G}(\mathbf{s}_L^T)||}_2 + {||\mathcal{P}^+-\mathbf{G}(\mathbf{s}^{T,\mathbf{\Delta}}_{L})||}_2
    \vspace{-0.2cm}
\end{equation}

To sum up, our overall training objective becomes: $\mathcal{L}_{\text{total}}$ = $\lambda_1\mathcal{L}_{\text{trip}}$ + $\lambda_2\mathcal{L}_{\text{comp}}$ + $\lambda_3\mathcal{L}_{\text{reg}}$ + $\lambda_4\mathcal{L}_{\text{TT}}$ + $\lambda_5\mathcal{L}_{\text{RT}}$ + $\lambda_6\mathcal{L}_{\text{rec}}$. Our model employs multitask learning with multiple losses updating the $\mathtt{LayerNorm}$ of $\mathbf{V}$, $\mathbf{C_{V2W}}$, $\mathbf{P}^L$, and $\mathbf{G}$. During inference, we discard the UNet decoder and use frozen $\mathbf{V}$, $\mathbf{T_{t}}$, and $\mathbf{C_{v2w}}$ to first generate a pseudo-word token from query sketch as: $\mathbf{s}^w=\mathbf{C_{v2w}(\mathbf{V}(\mathcal{S}))}$. We prepend $\mathbf{P}^L$ to $\mathbf{s}^w$ followed by appending the tokenised representation of the additional user-given textual description $\mathbf{t}^w = \mathbf{T_w}(\mathcal{T})$ to form the final composed query token $\{\mathbf{P}^L,\mathbf{s}^w, \mathbf{t}^w\}$. This composed query token, upon passing through $\mathbf{T_{t}}$, generates the final composed query feature $\mathbf{s}_L^{T,q}$. With pre-computed visual features of the gallery images (retrieval candidates), we perform retrieval by comparing the distance between the gallery features and $\mathbf{s}_L^{T,q}$ (both $l_2$ normalised).

\begin{figure*}[!htbp]
    \vspace{-0.2cm}
    \centering
    \includegraphics[width=\textwidth]{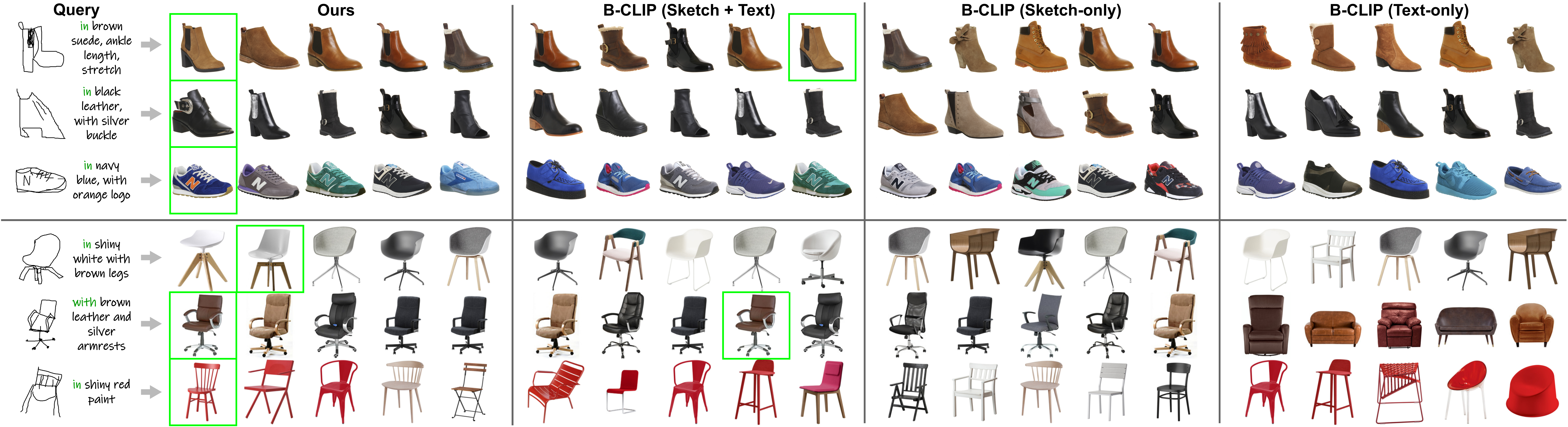}
    \vspace{-0.6cm}
    \caption{Top-5 \textit{fine-grained retrieval} result comparison on ShoeV2/ChairV2. GT photos are \GTgreen{green}-bordered. \textit{(Zoom-in for best-view)}}
    \label{fig:finegrained}
    \vspace{-0.4cm}
\end{figure*}

\vspace{-0.2cm}
\section{Experiments}
\vspace{-0.2cm}
\keypoint{Datasets.} We evaluate on the following datasets. QMUL-ShoeV2~\cite{yu2016sketch} and QMUL-ChairV2~\cite{yu2016sketch} contain $2000/6730$, and $400/1800$ sketches/photos respectively, with fine-grained association. The Sketchy~\cite{sangkloy2016the} dataset comprises $12,500$ photos across $125$ classes with at least $5$ sketches per photo. For scene-level retrieval, we use FS-COCO~\cite{chowdhury2022fs} and SketchyCOCO~\cite{gao2020sketchycoco} containing $10,000$ and $14,081$ paired sketch-text-photo triplets respectively, where images and textual captions are sourced from MS-COCO~\cite{lin2014microsoft}. We also use the ImageNet-R(endition)~\cite{hendrycks2021many} image-only dataset, which consists $30,000$ images across $200$ ImageNet~\cite{deng2009imagenet} classes and $16$ domains with domain annotations.

\keypoint{Implementation Details.} We use a pre-trained ViT-L/14 CLIP vision encoder~\cite{radford2021learning} in all experiments with an embedding dimension $d=768$. The prompt vectors are trained with a learning rate of $10^{-5}$, keeping the encoder frozen (except $\mathtt{LayerNorm}$ layers). The UNet decoder and the visual-to-word converter are trained with a learning rate of $10^{-4}$ and $10^{-3}$ respectively. We train the model for $100$ epochs using AdamW~\cite{loshchilov2019decoupled} optimiser with $0.09$ weight decay, and a batch size of $128$. Values of $\lambda_{1,2,3,4,5,6}$ are set to $1$, $0.5$, $0.1$, $0.1$, $1$, and $1$, empirically.

\keypoint{Competitors.} We evaluate from three perspectives --
\textit{(i) Sketch or Text-only \textbf{\underline{B}}aselines:} Here we validate the additional accuracy gain achieved by sketch+text composition over well-studied individual sketch/text-only retrieval paradigms. For sketch-only baselines, we use triplet loss-based frameworks~\cite{sain2023clip} of \textbf{B-DINOv2 (S)} and \textbf{B-CLIP (S)}, using DINOv2~\cite{oquab2023dinov2} and CLIP~\cite{radford2021learning} ViT-L/14 as backbones respectively. Given the de facto usage of
pre-trained CLIP~\cite{radford2021learning} in TBIR, for \textit{text-only} baselines we introduce \textbf{B-CLIP (T)}, which employs frozen CLIP~\cite{radford2021learning} vision and language encoder to retrieve by comparing query text feature against gallery image features. \textit{(ii) Sketch+Text Composed SoTA:} To judge CLIP's off-the-shelf potential in sketch+text composition, we employ \textbf{B-CLIP (S+T)} which uses the mean of sketch and text features from CLIP's frozen vision and language encoders for retrieval. Now, as most supervised sketch+text SoTA models~\cite{baldrati2022effective, sangkloy2022sketch, chowdhury2023scenetrilogy} train using \textit{paired textual captions}, we employ SoTA image captioner BLIP~\cite{li2023blip} to generate textual captions for training images which however are often noisy, generic and non-discriminative~\cite{li2023blip}. Among supervised SoTAs, \textbf{Combiner}~\cite{baldrati2022effective} trains a network to fuse paired image-text features of \textit{frozen} CLIP encoders~\cite{radford2021learning} to generate multi-modal query features, while \textbf{TASK-former}~\cite{sangkloy2022sketch} merges paired sketch and text features via element-wise addition and \textit{trains} CLIP's vision and language encoders~\cite{radford2021learning} end-to-end. \textbf{SceneTrilogy}~\cite{chowdhury2023scenetrilogy} models sketch-text-photo joint-embedding by training an invertible neural network.
\textit{(iii) Unsupervised Sketch+Text Composition:} Leveraging \textit{unlabelled photos}, \textbf{Pic2Word}~\cite{saito2023pic2word} learns a textual-inversion network to map input visual query into a pseudo-word token in CLIP's textual embedding space for retrieval. Unlike Pic2Word, \textbf{SEARLE}~\cite{baldrati2023zero} generates a set of pseudo-word tokens from \textit{unlabelled photos} using optimisation-based textual-inversion, and then uses those image-token pairs to learn a textual-inversion network. While such methods either train a textual-inversion network on a massive $3M$ image dataset~\cite{saito2023pic2word} or use time-consuming optimisation-based textual-inversion for image-token pair generation~\cite{baldrati2023zero}, we exploit pre-trained CLIP model to address sketch+text compositionality in an \textit{unsupervised} manner without any associated textual descriptions. We adapt Combiner~\cite{baldrati2022effective}, Pic2Word~\cite{saito2023pic2word}, and SEARLE~\cite{baldrati2023zero}, by replacing the input \textit{image} with \textit{sketch}. For fairness, we keep the same training/testing paradigm for all competing methods.

\begin{table*}[!htbp]
\vspace{-0.3cm}
\setlength{\tabcolsep}{10pt}
\renewcommand{\arraystretch}{0.8}
\centering
\caption{Results for \emph{fine-grained} object-level and scene-level composed retrieval.}
\vspace{-0.3cm}
\label{tab:fgsbir}
\notsotiny
\begin{tabular}{lcccccc|cccc}
\toprule
\multicolumn{1}{c}{\multirow{3}{*}{Methods}} & \multicolumn{6}{c|}{Object-level} & \multicolumn{4}{c}{Scene-level} \\ \cmidrule(lr){2-7}\cmidrule(lr){8-11}
 & \multicolumn{2}{c}{ShoeV2} & \multicolumn{2}{c}{ChairV2} & \multicolumn{2}{c|}{Sketchy} & \multicolumn{2}{c}{FS-COCO}  & \multicolumn{2}{c}{SketchyCOCO}  \\
\cmidrule(lr){2-3}\cmidrule(lr){4-5}\cmidrule(lr){6-7}\cmidrule(lr){8-9}\cmidrule(lr){10-11}
& Acc.@5 & Acc.@10 & Acc.@5 & Acc.@10 & Acc.@5 & Acc.@10 & Acc.@5 & Acc.@10 & Acc.@5 & Acc.@10  \\
\cmidrule(lr){1-3}\cmidrule(lr){4-5}\cmidrule(lr){6-7}\cmidrule(lr){8-9}\cmidrule(lr){10-11}
B-CLIP (S)             & 9.8  & 17.5 & 16.7 & 18.4 & 6.8 & 11.3 & 5.9 & 9.7 & 6.8 & 10.2 \\
B-DINOv2 (S)           & 10.2  & 19.4 & 17.9 & 20.2 & 8.5 & 15.1 & 7.6 & 11.4 & 9.4 & 12.2 \\
B-CLIP (T)             & 9.1 & 16.6 & 15.4 & 17.8 & 5.9 & 10.2 & 5.5 & 10.1 & 6.7 & 10.6 \\
\cmidrule(lr){1-3}\cmidrule(lr){4-5}\cmidrule(lr){6-7}\cmidrule(lr){8-9}\cmidrule(lr){10-11}
B-CLIP (S+T)           & 19.1  & 30.8 & 30.2 & 32.3 & 10.1 & 20.2 & 10.2 & 15.4 & 11.2 & 20.2 \\
Combiner~\cite{baldrati2022effective}  & 24.7  & 40.2 & 35.7 & 39.9 & 15.7 & 33.7 & 11.6 & 22.1 & 15.9 & 32.2 \\
TASK-former~\cite{sangkloy2022sketch}            & 27.7  & 44.1 & 40.7 & 45.2 & 17.8 & 35.2 & 12.7 & 24.2 & 19.4 & 34.7 \\
SceneTrilogy~\cite{chowdhury2023scenetrilogy}            & 29.1  & 46.2 & 43.4 & 46.8 & 19.7 & 37.2 & 14.5 & 28.3 & 20.4 & 40.2 \\
\cmidrule(lr){1-3}\cmidrule(lr){4-5}\cmidrule(lr){6-7}\cmidrule(lr){8-9}\cmidrule(lr){10-11}
Pic2Word~\cite{saito2023pic2word}     & 34.7  & 58.4 & 55.7 & 62.1 & 22.5 & 48.7 & 16.7 & 32.6 & 24.4 & 46.0 \\
SEARLE~\cite{baldrati2023zero}        & 38.4  & 64.8 & 60.8 & 66.4 & 25.3 & 54.2 & 17.7 & 35.9 & 26.0 & 50.4 \\
\rowcolor{YellowGreen!40}
\textbf{\textit{Proposed}}            & \bf47.3  & \bf79.1 & \bf73.5 & \bf81.4 & \bf30.6  & \bf64.2 & \bf22.7 & \bf43.5 & \bf33.4 & \bf61.1 \\
\textit{\graytext{Avg. Improvement}}   & \graytext{\textit{+24.7}}  & \graytext{\textit{+45.5}} & \graytext{\textit{+38.3}} & \graytext{\textit{+42.6}} & \graytext{\textit{+15.9}} & \graytext{\textit{+34.6}} & \graytext{\textit{+11.3}} & \graytext{\textit{+22.4}} & \graytext{\textit{+17.8}} & \graytext{\textit{+32.5}} \\
\bottomrule
\end{tabular}
\vspace{-0.35cm}
\end{table*}

\keypoint{Evaluation Setup.} In the fine-grained setup, we aim to retrieve the target image using the composed query formed by an input sketch and text. We use a train:test split of $90$:$10$ for Sketchy \cite{sangkloy2016the} (following \cite{sain2023exploiting}), $7000$:$3000$ for FS-COCO \cite{chowdhury2022fs} and $1015$:$210$ for SketchyCOCO \cite{gao2020sketchycoco}. For ShoeV2/ChairV2 we use $1800/300$ ($6051/1275$) photos (sketches) for training and the rest for testing. Notably, our method \textit{does not} use captions during training. For \textit{evaluation}, we manually collect fine-grained captions for each of the test-set images of ShoeV2, ChairV2, and Sketchy. We compose the query as $\{\mathbf{P}^L;\mathbf{s}^w;\mathtt{[text]}\}$, where $\mathtt{[text]}$ denotes the word embedding of textual query with suitable prepositions (\eg, `with', `in'). We use Acc.@q to denote the percentage of sketches with true-matched photos in the top-q retrieved images.

\subsection{Performance Analysis}
\vspace{-0.2cm}
\cref{tab:fgsbir} delineates the quantitative results while the qualitative ones are shown in \cref{fig:finegrained}. In the fine-grained composed retrieval setup (\cref{tab:fgsbir}), our method outperforms baselines and SoTAs significantly on all datasets, indicating its efficiency in seamlessly combining fine-grained sketch with textual description. This gain is likely due to the regularisation provided by our region-aware contrastive loss and generator guidance. Our closest competitors (\ie, Pic2Word \cite{saito2023pic2word}, and SEARLE \cite{baldrati2023zero}) attempt to handle composed retrieval by either training data-hungry textual-inversion network~\cite{saito2023pic2word} or use optimisation-based textual-inversion for image-token pair generation followed by training an inversion network \cite{baldrati2023zero}. Surprisingly, our method achieves the highest Acc.@5 of $47.3~(73.5)$ in ShoeV2 (ChairV2) without the $3M$ data-requirement of Pic2Word \cite{saito2023pic2word}, or the complicated two-stage approach of SEARLE \cite{baldrati2023zero}.

Being more challenging than object-level \cite{chowdhury2022partially}, baseline methods perform quite poorly (\cref{tab:fgsbir}) for scene-image retrieval. However, thanks to the increased interaction capability (learned via compositionality constraint) of the pseudo-word token with user-given textual queries during inference, we surpass others with an Acc.@5 of $22.7~(33.4)$ on FS-COCO (SketchyCOCO).

\vspace{-0.1cm}
\subsection{Downstream Tasks}
\vspace{-0.2cm}

\begin{figure}[!htbp]
    \centering
    \includegraphics[width=\columnwidth]{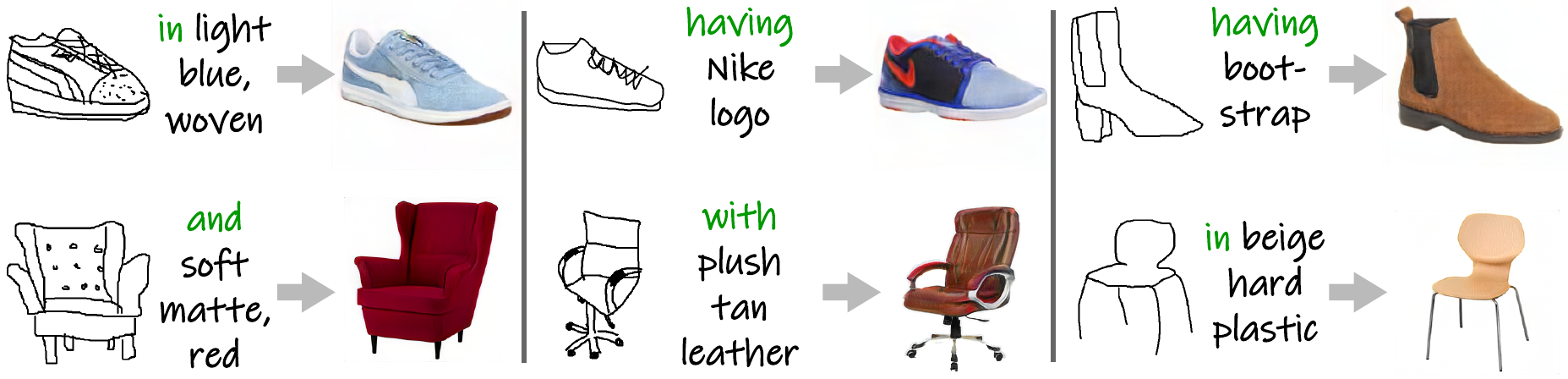}
    \vspace{-0.6cm}
    \caption{Qualitative results for \textit{sketch+text composed fine-grained generation} with pre-trained StyleGAN2~\cite{karras2020analyzing} models.}
    \label{fig:generation}
    \vspace{-0.6cm}
\end{figure}

\keypoint{Sketch+Text-based Fine-Grained Image Generation.} Apart from composed retrieval, our method is also suitable for sketch+text composed object image generation. Here we replace the low-quality UNet decoder with a StyleGAN2~\cite{karras2020analyzing} generator (pre-trained on specific classes). The composed query $\mathbf{s}_L^{T,q}$ (acting as a latent vector \cite{karras2020analyzing}) upon passing through the frozen StyleGAN2 generates output photos. We pass $\mathbf{s}_L^{T,q}$ via a learnable FC-layers to convert it to the dimensions required by StyleGAN2's affine transformation layer~\cite{karras2020analyzing}. \cref{fig:generation} shows a few cases of such fine-grained generation. Notably, the \textit{semantic geometry} (\eg, shape, structure, etc.) of generated images is driven by input sketches, whereas the high-level appearance (\eg, colour, shade, etc.) is mostly governed by textual descriptions. Overall our method archives a lower FID~\cite{karras2020analyzing, koley2023its} score of $33.4 (88.5)$ on ShoeV2 (ChairV2) test-set images compared to $35.85 (90.21)$ of the current SoTA~\cite{koley2023picture}.

\vspace{-0.2cm}

\begin{figure}[!htbp]
    \vspace{-0.32cm}
    \centering
    \includegraphics[width=\columnwidth]{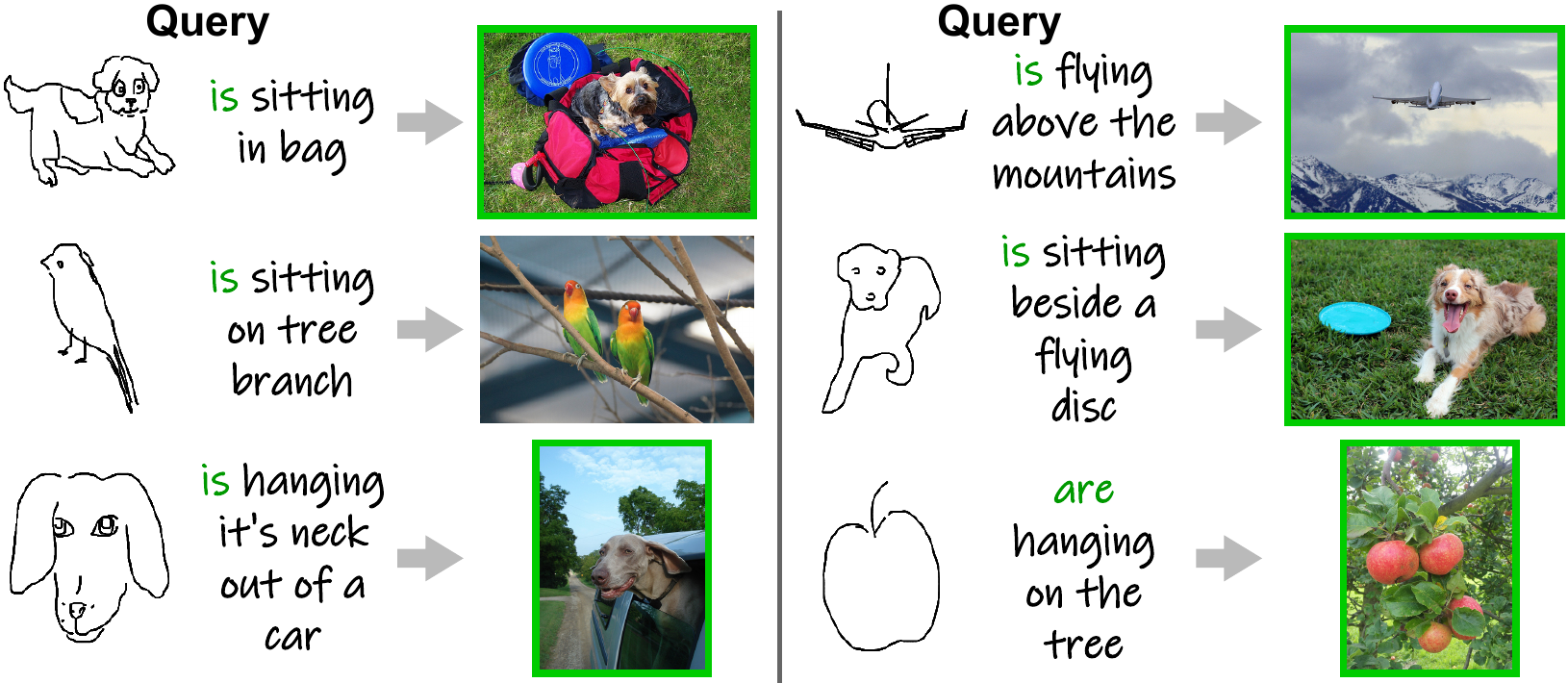}
    \vspace{-0.6cm}
    \caption{Qualitative result for \textit{object sketch-based scene image retrieval} on FS-COCO~\cite{chowdhury2022fs}. GT photos are \GTgreen{green}-bordered.}
    \label{fig:scene}
    \vspace{-0.4cm}
\end{figure}

\keypoint{Object Sketch-based Scene Image Retrieval.} It aims to retrieve \textit{scene} images from a \textit{single-object} sketch and additional captions. Here, we use $7000(3000)$ and $1015(210)$ train(test) sketch-photo pairs from FS-COCO and SketchyCOCO respectively. Since they lack single-object sketches, we source them from Sketchy, which is their superset in terms of classes \cite{chowdhury2022fs}. Here, a retrieval is deemed correct if it contains \textit{all} objects that were queried in the sketch and text. As FS-COCO/SketchyCOCO images are derived from MS-COCO \cite{lin2014microsoft}, we use their segmentation-map \textit{labels} from MS-COCO to create ground truth object-lists per test-set image. During inference, we use the readily available captions of test-set images of FS-COCO/SketchyCOCO, but remove the object-\textit{name} queried via sketch (\cref{fig:scene}). Here we compose the query like in the fine-grained composed retrieval setup and use Acc.@q as evaluation metric. \cref{tab:domain} shows our method to surpass other baseline and SoTAs with an average Acc.@5 gain of $10.9$ on FS-COCO \cite{chowdhury2022fs}.

\keypoint{Domain Attribute Transfer.} Here we perform SBIR from a specified domain (\eg, origami), where the domain name is additionally provided by a textual domain label. Here we use ImageNet-R \cite{hendrycks2021many} dataset. This being an image-only dataset, we source freehand sketches from the $104$ common classes from the Sketchy \cite{sangkloy2016the} dataset with a train:test split of $90$:$10$. Here, a retrieval is deemed correct, if its class and domain name match those of the query sketch and domain label. Due to non-uniform and noisy domain labels, we use four domains here \textit{viz.}, tattoo, origami, sculpture, and painting. We compose the query as $\{\mathbf{P}^L;\mathbf{s}^w;\mathtt{[in]};\mathtt{[domain]}\}$, where $\mathtt{[domain]}$ denotes the word embedding of the query domain label. Following \cite{saito2023pic2word}, we use recall@q (r@q) as the evaluation metric, which denotes the ratio of positive retrieved images in the top-q list to all relevant images for a given query. As strict domain constraints complicate cross-modal composed image retrieval, baseline methods perform poorly here, with B-CLIP (S+T) attaining an average r@50 of $12.9$ across four test domains (\cref{tab:domain}). Due to their respective CLIP \cite{radford2021learning} feature composition strategies, Combiner \cite{baldrati2022effective}, Pic2Word \cite{saito2023pic2word} and SEARLE \cite{baldrati2023zero} depict reasonable performance on ImageNet-R, while our method achieves a notable average r@10 of $15.3$.

\begin{figure}[!htbp]
    \vspace{-0.3cm}
    \centering
    \includegraphics[width=\columnwidth]{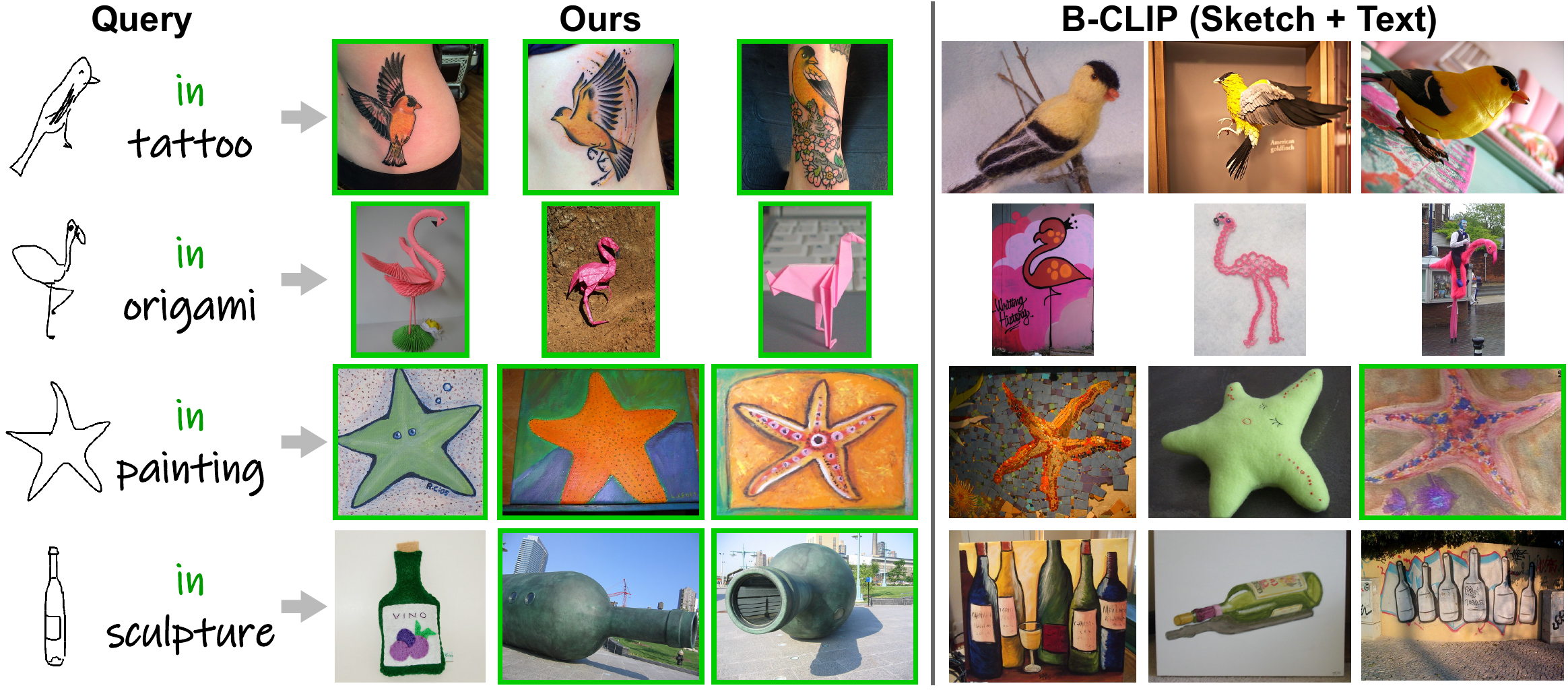}
    \vspace{-0.75cm}
    \caption{Top-3 \textit{domain attribute transfer} results comparison on ImageNet-R~\cite{hendrycks2021many}. GT photos are \GTgreen{green}-bordered.}
    \vspace{-0.6cm}
    \label{fig:domain}
\end{figure}

\begin{table}[!htbp]
\setlength{\tabcolsep}{3pt}
\renewcommand{\arraystretch}{1}
\centering
\caption{Results for \textit{domain attribute transfer} and \textit{object sketch-based scene image retrieval}.}
\vspace{-0.3cm}
\label{tab:domain}
\scriptsize
    \begin{tabular}{lcc|cccc}
    \toprule
        \multicolumn{1}{c}{\multirow{3}{*}{Methods}}   & \multicolumn{2}{c|}{Domain transfer} & \multicolumn{4}{c}{Object sketch-based scene retrieval}\\\cmidrule(lr){2-3}\cmidrule(lr){4-7}
            & \multicolumn{2}{c|}{ImageNet-R} & \multicolumn{2}{c}{FS-COCO} & \multicolumn{2}{c}{SketchyCOCO} \\\cmidrule(lr){2-3}\cmidrule(lr){4-7}
                                                        & r@10  & r@50  & Acc.@5 & Acc.@10 & Acc.@5 & Acc.@10    \\
         \cmidrule(lr){1-7}
         B-CLIP (S+T)                                  & 2.2 & 12.9 & 4.3 & 7.0 & 10.1 & 26.7  \\
         Combiner~\cite{baldrati2022effective}          & 8.3 & 19.8 & 10.8 & 21.1 & 16.7 & 30.9 \\
         TASK-former~\cite{sangkloy2022sketch}          & 7.6 & 18.2 & 11.6 & 23.1 & 18.2 & 35.8 \\
         SceneTrilogy~\cite{chowdhury2023scenetrilogy}  & 9.7 & 20.5 & 14.2 & 25.3 & 21.0 & 40.8 \\
         \cmidrule(lr){1-7}
         Pic2Word~\cite{saito2023pic2word}              & 10.8 & 22.7 & -- & -- & -- & -- \\
         SEARLE~\cite{baldrati2023zero}                 & 12.1 & 25.4 & -- & -- & -- & -- \\
         \rowcolor{YellowGreen!40}
         \textbf{\textit{Proposed}}                     & \bf15.3 & \bf27.1 & \bf21.2 & \bf40.4 & \bf32.4 & \bf53.3 \\
         \textit{\graytext{Avg. Improvement}}           & \graytext{\textit{+6.8}}  & \graytext{\textit{+7.1}} & \graytext{\textit{+10.9}} & \graytext{\textit{+21.2}} & \graytext{\textit{+15.9}} & \graytext{\textit{+19.4}} \\
    \bottomrule
    \end{tabular}
    \vspace{-0.6cm}
\end{table}

\begin{table}[!htbp]
\setlength{\tabcolsep}{2pt}
\renewcommand{\arraystretch}{1}
\centering
\caption{Ablation on design.}
\vspace{-0.3cm}
\label{tab:abal}
\scriptsize
\begin{tabular}{lccccccc}
\toprule
\multicolumn{1}{c}{\multirow{2}{*}{Methods}} & \multicolumn{2}{c}{ShoeV2} & \multicolumn{2}{c}{ChairV2} & \multicolumn{2}{c}{FS-COCO} \\
\cmidrule(lr){2-3}\cmidrule(lr){4-5}\cmidrule(lr){6-7}
                                            & Acc.@5 & Acc.@10 & Acc.@5 & Acc.@10 & Acc.@5 & Acc.@10 \\
\cmidrule(lr){1-3}\cmidrule(lr){4-5}\cmidrule(lr){6-7}
w/o $\mathcal{L}_{\text{TT}}$               & 41.8  & 71.9 & 68.3 & 77.7 & 17.1 & 38.2 \\
w/o $\mathcal{L}_{\text{rec}}$              & 40.7  & 72.8 & 70.9 & 75.5 & 21.4 & 42.7 \\
w/o $\mathcal{L}_{\text{RT}}$               & 40.2  & 71.6 & 69.1 & 76.2 & 10.5 & 21.7 \\
w/o compositionality                        & 32.5  & 48.2 & 45.7 & 48.9 & 18.4 & 32.3 \\
\rowcolor{YellowGreen!40}
\textbf{\textit{Ours-full}}                 & \bf47.3  & \bf79.1 & \bf73.5 & \bf81.4 & \bf22.7 & \bf43.5\\
\bottomrule
\end{tabular}
\vspace{-0.2cm}
\end{table}

\subsection{Ablation on Design}
\begin{figure}[!htbp]
    \centering
    \includegraphics[width=1\linewidth]{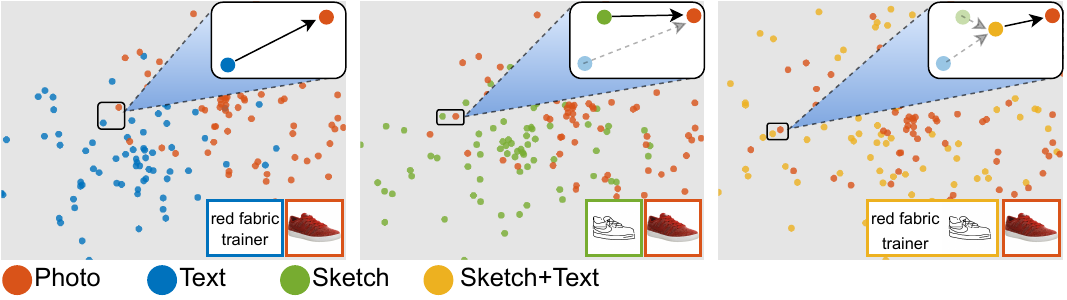}
    \vspace{-0.6cm}
    \caption{t-SNE plots showing the feature distances for text-based, sketch-based, and composed retrieval. Compared to sketch/text-based retrieval, combining sketch and text pushes the composed embedding closer to the ground truth photo in the latent manifold.}
    \label{fig:tsne}
    \vspace{-0.5cm}
\end{figure}

\vspace{-0.1cm}
\noindent$\bullet$ \textbf{How well does $\mathbf{s}^w$ capture sketch semantics?} To judge the efficacy of pseudo-word token $\mathbf{s}^w$ in representing visual content of a sketch, we evaluate our trained model on retrieving an input sketch \textit{solely} from its pseudo-word token \textit{without} additional textual description. Acc.@1 of $98.35\%$ on ShoeV2 \textit{test-set} sketches shows the pseudo-word token to capture visual sketch features fairly well. Although representing fine-grained query sketches with one pseudo-word token might be sub-optimal, experimenting with two and three such tokens delivered similar Acc.@1 of $98.79\%$ and $99.11\%$ respectively on ShoeV2. We thus stick to a single-word token for computational ease.\\
$\bullet$ \textbf{Contribution of $\mathcal{L}_{\text{RT}}$ and $\mathcal{L}_{\text{rec}}$:} Region-aware local features are pivotal in bridging the huge domain gap between sparse-binary sketches and pixel-dense photos. Although less reflected on ShoeV2/ChairV2 results, a notable Acc.@5 drop of $12.2$ on FS-COCO (\cref{tab:abal}) for \textbf{w/o $\mathcal{L}_{\text{RT}}$} verifies its importance in the scene-level setup. Furthermore, a $13.9\%$ drop in Acc.@5 (ShoeV2) for \textbf{w/o $\mathcal{L}_{\text{rec}}$} shows that fine-grained matching remains incomplete without the proposed generator guidance. \\
$\bullet$ \textbf{Impact of $\mathcal{L}_{\text{TT}}$:}
Removing Text-to-Text loss plummets Acc.@5 by $11.6\%$ on ShoeV2 (\cref{tab:abal}), highlighting its vital role in aligning learned and English language prompts seen by CLIP~\cite{radford2021learning} during its training. This removal likely pushes the final language embedding towards sparser parts of CLIP language manifold~\cite{baldrati2023zero}, hindering effective communication with \textit{real-language} tokens during inference~\cite{baldrati2023zero}.\\
$\bullet$ \textbf{Why Compositionality Constraint?} Introduced via the novel idea of \textit{neutral text}, compositionality constraint helps preserve internal grammar of CLIP language manifold, to allow optional user-provided query texts during inference. On removing that, Acc.@5 drops the lowest across all datasets (\cref{tab:abal}), proving its importance in our framework.\\
$\bullet$ \textbf{On combining sketch and text:} Our composed retrieval pipeline places the query feature \textit{closer} to its paired photo in the latent retrieval space (\cref{fig:tsne}), than \textit{only} text/sketch-based retrieval. The relative distances (\cref{fig:tsne} top insets) depict that the individual text and sketch feature together \textit{pushes} the composed feature towards the paired photo.

\vspace{-0.2cm}

\vspace{-0.2cm}
\section{Conclusion and Future Works}
\vspace{-0.1cm}

In conclusion, our exploration into the fine-grained representation power of both sketch and text, coupled with the orchestration of their synergistic interplay, marks a significant stride in the realm of image retrieval. By harmonising sketches and text, we offer users a retrieval experience that transcends traditional category-level distinctions. The introduction of a novel compositionality framework, driven by pre-trained CLIP models, eliminates the need for extensive fine-grained textual annotations. Last but not least, our system extends its utility to diverse domains such as sketch+text-based fine-grained image generation, object-sketch-based scene retrieval, domain attribute transfer, etc.

{
    \small
    \bibliographystyle{ieeenat_fullname}
    \bibliography{arxiv}
}

\clearpage

\twocolumn[{\centering{\Large \textbf{Supplementary material for\\ You'll Never Walk Alone: A Sketch and Text Duet for \\ Fine-Grained Image Retrieval}\par}\vspace{0.3cm}
	{\MYhref[cvprblue]{https://subhadeepkoley.github.io}{Subhadeep Koley}\textsuperscript{1,2} \hspace{.2cm} \MYhref[cvprblue]{https://ayankumarbhunia.github.io}{Ayan Kumar Bhunia}\textsuperscript{1} \hspace{.2cm} \MYhref[cvprblue]{https://aneeshan95.github.io}{Aneeshan Sain}\textsuperscript{1} \hspace{.2cm}  \MYhref[cvprblue]{https://www.pinakinathc.me}{Pinaki Nath Chowdhury}\textsuperscript{1}\\ \MYhref[cvprblue]{https://www.surrey.ac.uk/people/tao-xiang}{Tao Xiang}\textsuperscript{1,2} \hspace{.2cm} \MYhref[cvprblue]{https://www.surrey.ac.uk/people/yi-zhe-song}{Yi-Zhe Song}\textsuperscript{1,2} \\
\textsuperscript{1}SketchX, CVSSP, University of Surrey, United Kingdom.  \\
\textsuperscript{2}iFlyTek-Surrey Joint Research Centre on Artificial Intelligence.\vspace{0.1cm}\\
{\tt\small \{s.koley, a.bhunia, a.sain, p.chowdhury, t.xiang, y.song\}@surrey.ac.uk}\par\vspace{0.5cm}}}
]

\section*{A. Composed Fashion Image Retrieval}
\begin{figure}[!htbp]
    \centering
    \includegraphics[width=1\linewidth]{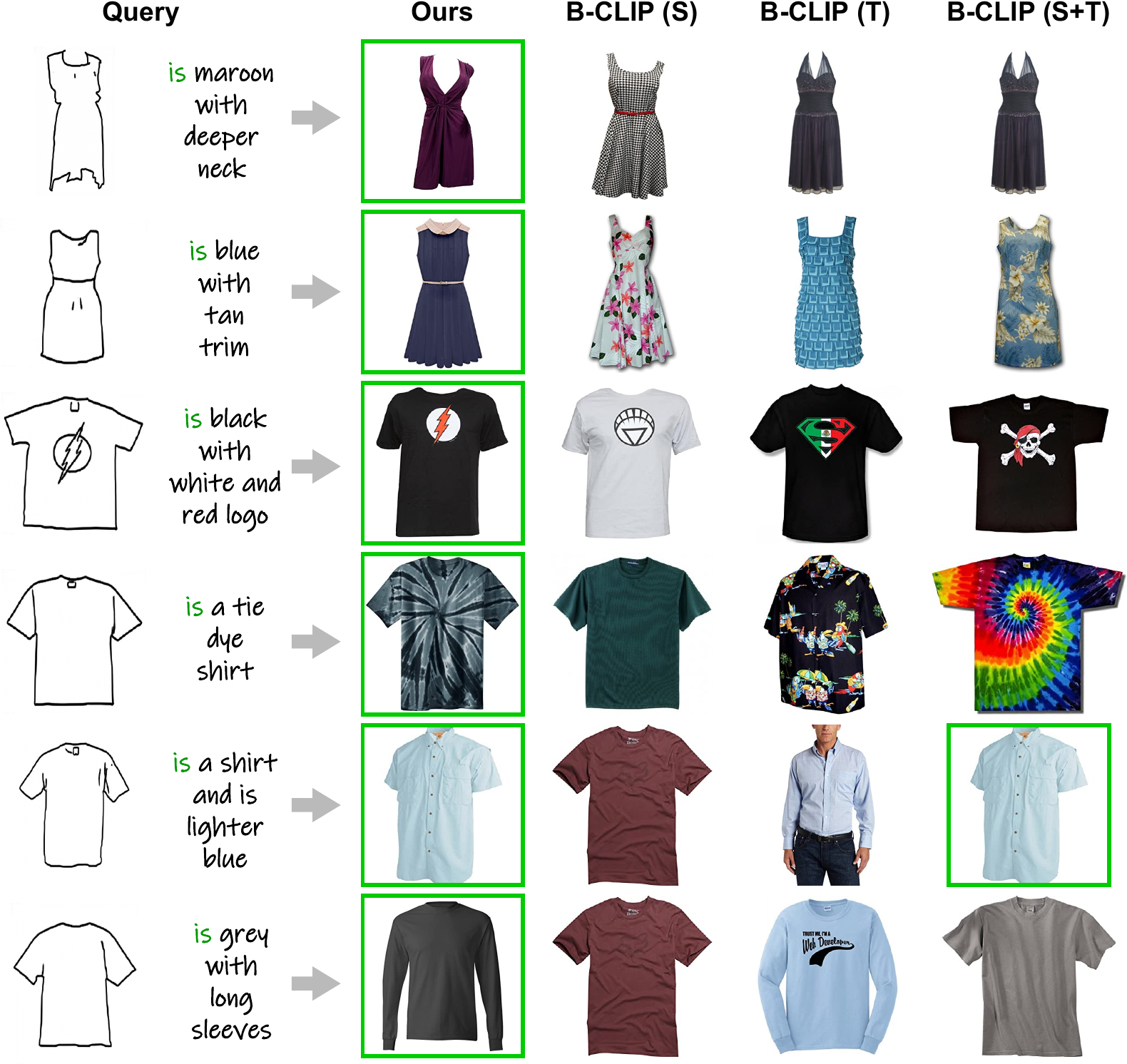}
    \vspace{-0.4cm}
    \caption{Qualitative comparison with baselines for sketch+text composed fashion image retrieval on FashionIQ~\cite{wu2021fashion}. GT photos are \GTgreen{green}-bordered. Notably, even though the images retrieved by B-Sketch+Text are mostly of the same shape as the query sketch, they lack the desired appearance given by textual description.}
    \label{fig:fashion}
    \vspace{-0.4cm}
\end{figure}

Although our method is primarily focused on \textit{fine-grained} composed image retrieval, here we showcase an interesting application of our method namely, \textit{sketch+text composed fashion image retrieval}. It aims to retrieve fashion garment images from abstract sketches and additional textual queries. We train and test our method on FashionIQ~\cite{wu2021fashion} dataset, having $\sim$$15$K paired triplets of query-image, textual description and target-image. Due to the lack of paired sketches, we generate synthetic sketches from query images using \cite{chan2022drawings}. Compared to other baselines (\cref{fig:fashion}), images retrieved by the proposed method hold more semantic coherence with the sketch+text composed query.

\vspace{-0.2cm}

\section*{B. Details on Neutral Text Set}
We aim to restrict the adverse effects of sketch-photo difference signal $\mathbf{\Delta}$ to a minimum. For that, we regularise the training via a \textit{``neutral-text''} set containing a list of $3$-$5$ word \emph{generic} descriptions of a freehand abstract sketch (\cref{composition}), generated via a lightweight GPT \cite{brown2020language}. The full neutral-text set is given below:
\\
\\
\noindent\texttt{"in abstract lines",}\\
\texttt{"as sparse contours",}\\
\texttt{"with a line drawing",}\\
\texttt{"as an artwork",}\\
\texttt{"as a ballpoint sketch",}\\
\texttt{"in chirography",}\\
\texttt{"as a blueprint",}\\
\texttt{"in cartography",}\\
\texttt{"as a cartoon",}\\
\texttt{"in charcoal sketch",}\\
\texttt{"as conceptual drawings",}\\
\texttt{"in design sketches",}\\
\texttt{"as concept sketches",}\\
\texttt{"in professional sketches",}\\
\texttt{"as freehand sketches",}\\
\texttt{"as contour maps",}\\
\texttt{"as edge maps",}\\
\texttt{"of pencil curves",}\\
\texttt{"as a sparse diagram",}\\
\texttt{"as doodles",}\\
\texttt{"in drawing",}\\
\texttt{"in ink etching",}\\
\texttt{"as a deformed figure",}\\
\texttt{"as a line diagram",}\\
\texttt{"in geometrical drawing",}\\
\texttt{"containing sparse strokes",}\\
\texttt{"as graphical representations",}\\
\texttt{"containing hatching",}\\
\texttt{"as an illustration",}\\
\texttt{"as a sparse imitation",}\\
\texttt{"with sketch impression",}\\
\texttt{"as an ink sketch",}\\
\texttt{"as an ink doodle",}\\
\texttt{"as an ink drawing",}\\
\texttt{"in graphical layout",}\\
\texttt{"as line drawing",}\\
\texttt{"as line art",}\\
\texttt{"as line sketch",}\\
\texttt{"in chirographic representation",}\\
\texttt{"as monochrome drawing",}\\
\texttt{"with black and white outlines",}\\
\texttt{"in sketchy pattern",}\\
\texttt{"as a pencil drawing",}\\
\texttt{"as a pencil sketch",}\\
\texttt{"as a pencil doodle",}\\
\texttt{"containing pencil strokes",}\\
\texttt{"as perspective drawing",}\\
\texttt{"with hatching representation",}\\
\texttt{"as a contour plot",}\\
\texttt{"in sketch portrayal",}\\
\texttt{"as preliminary drawing",}\\
\texttt{"as preliminary sketch",}\\
\texttt{"in stroke rendering",}\\
\texttt{"as doodle representation",}\\
\texttt{"as rough draft",}\\
\texttt{"as rough drawing",}\\
\texttt{"as a rough outline",}\\
\texttt{"in contour schematic",}\\
\texttt{"as a scribble",}\\
\texttt{"as a rough shape",}\\
\texttt{"in silhouette",}\\
\texttt{"as a skeleton drawing",}\\
\texttt{"as a stipple",}\\
\texttt{"containing line traces",}\\
\texttt{"as a line tracing",}\\
\texttt{"as rough doodle",}\\
\texttt{"with artistic rendering",}\\
\texttt{"as a graphite sketch",}\\
\texttt{"as a freeform drawing",}\\
\texttt{"as a quick sketch",}\\
\texttt{"with pen and ink drawing",}\\
\texttt{"as an unconstrained sketch",}\\
\texttt{"as a rapid sketch",}\\
\texttt{"as an impromptu drawing",}\\
\texttt{"having loose pen strokes",}\\
\texttt{"as a rough sketch",}\\
\texttt{"as a gesture drawing",}\\
\texttt{"as a cartoon",}\\
\texttt{"as an abstract sketch",}\\
\texttt{"as a thumbnail sketch",}\\
\texttt{"as a hand drawn illustration",}\\
\texttt{"as fine art",}\\
\texttt{"as imaginary sketch",}\\
\texttt{"as casual drawing",}\\
\texttt{"as spontaneous sketch",}\\
\texttt{"as organic drawing",}\\
\texttt{"as experimental drawing",}\\
\texttt{"as an unstructured sketch",}\\
\texttt{"as a naturalistic drawing",}\\
\texttt{"as a contour drawing",}\\
\texttt{"as expressive sketch",}\\
\texttt{"containing whimsical sketch strokes",}\\
\texttt{"as an artistic drawing",}\\
\texttt{"with uninhibited sketch strokes",}\\
\texttt{"as minimalist drawing",}\\
\texttt{"as a free-flowing drawing",}\\
\texttt{"as a symbolic sketch",}\\
\texttt{"having unrestricted strokes",}\\
\texttt{"as a playful doodle",}\\
\texttt{"with monochrome brush strokes",}\\

\section*{C. Details on Handcrafted Prompts}
Our \textit{text-to-text} generalisation loss (\cref{text_text}) enforces the learned prompts to be similar to handcrafted English language prompts, in the text embedding space for better generalisation. In particular, at every instance, we randomly pick one handcrafted fixed language prompt from the below set, which we curate from existing prompt learning literature~\cite{bulat2023lasp, radford2021learning, zhou2022conditional}.
\\
\\
\noindent\texttt{"a photo of",}\\
\texttt{"an image of",}\\
\texttt{"itap of a",}\\
\texttt{"a photo of the hard to see",}\\
\texttt{"a low resolution photo of the",}\\
\texttt{"a rendering of a",}\\
\texttt{"a bad photo of the",}\\
\texttt{"a photo of a person doing",}\\
\texttt{"a high-resolution photo of",}\\
\texttt{"an origami of",}\\
\texttt{"a cropped photo of the",}\\
\texttt{"a pixelated photo of the",}\\
\texttt{"a bright photo of the",}\\
\texttt{"a close-up photo of the",}\\
\texttt{"a low resolution photo of a",}\\
\texttt{"a rendition of the",}\\
\texttt{"a clear photo of the",}\\
\texttt{"a blurry photo of a",}\\
\texttt{"a pixelated photo of a",}\\
\texttt{"itap of the",}\\
\texttt{"a photo of the small",}\\
\texttt{"a photo of the large",}\\
\texttt{"a black and white photo of",}\\
\texttt{"an art of the",}\\
\texttt{"a photo of a",}\\
\texttt{"a photo of many",}\\
\texttt{"a cropped photo of a",}\\
\texttt{"a photo of the",}\\
\texttt{"a good photo of the",}\\
\texttt{"a rendering of the",}\\
\texttt{"a photo of a large",}\\
\texttt{"a jpeg corrupted photo of the",}\\
\texttt{"a good photo of a"}\\
\texttt{"a photograph of a"}\\
\texttt{"a 4K photo of"}\\
\texttt{"a photorealistic image of a"}\\
\texttt{"a snapshot of"}\\
\texttt{"a good picture of"}\\
\texttt{"a bad picture of"}\\
\texttt{"a sharp photograph of"}\\
\texttt{"a blurry photograph of"}\\

\vspace{-0.3cm}
\section*{D. Full List of Connecting Words}

In this paper, we represent an input sketch as a \textit{pseudo-word token} that emulates its visual concept in equivalent word-embedding space. Subsequently, we combine it with the textual description via connecting phrases to obtain ``$\langle$pseudo-word token$\rangle$ $\langle$connecting phrase$\rangle$ $\langle$text description$\rangle$" that forms our composed query. The full list of such connecting phrases is given below:
\\
\\
\noindent\texttt{is, in, having, are, and, with, upon, on, containing, comprising, of, beside, including, over, under, alongside, plus, without, above, below, as.}

\end{document}